\documentclass[sigconf]{acmart}

\copyrightyear{2026}
\acmYear{2026}
\setcopyright{cc}
\setcctype{by}
\acmConference[MM '26]{Proceedings of the 34th ACM International Conference on Multimedia}{November 10--14, 2026}{Rio de Janeiro, Brazil}
\acmBooktitle{Proceedings of the 34th ACM International Conference on Multimedia (MM '26), November 10--14, 2026, Rio de Janeiro, Brazil}
\acmDOI{10.1145/3767308.3835208}
\acmISBN{979-8-4007-2213-4/2026/11}

\usepackage{amsmath}
\usepackage{mathtools}
\usepackage{amsthm}
\usepackage{adjustbox}
\usepackage{multirow}

\begin{document}

\title{High-Fidelity Face Content Recovery via Tamper-Resilient Versatile Watermarking}

\author{Peipeng Yu}

\affiliation{%
  \institution{Jinan University}
  \city{Guangzhou}
  \state{Guangdong}
  \country{China}
}
\email{ypp865@163.com}

\author{Jinfeng Xie}

\affiliation{%
  \institution{Jinan University}
  \city{Guangzhou}
  \state{Guangdong}
  \country{China}}
\email{jinfengxie@stu2024.jnu.edu.cn}

\author{Chengfu Ou}

\affiliation{%
  \institution{Jinan University}
  \city{Guangzhou}
  \state{Guangdong}
  \country{China}
}
\email{hahally@stu2025.jnu.edu.cn}
\author{Xiaoyu Zhou}

\affiliation{%
  \institution{Jinan University}
  \city{Guangzhou}
  \state{Guangdong}
  \country{China}
}
\email{xiaoyuuzhou68@163.com}
\author{Jianwei Fei}

\affiliation{%
 \institution{University of Macau}
 \city{Macau}
 \country{China}}
\email{fei\_jianwei@163.com}
\author{Yunshu Dai}

\affiliation{%
  \institution{Sun Yat-sen University}
  \city{Guangzhou}
  \state{Guangdong}
  \country{China}}
\email{daiyunshu0102@163.com}
\author{Zhihua Xia}
\authornote{Corresponding author.}

\affiliation{%
  \institution{Jinan University}
  \city{Guangzhou}
  \state{Guangdong}
  \country{China}}
\email{xia\_zhihua@163.com}

\author{Chip Hong Chang}
\affiliation{%
  \institution{Nanyang Technological University}
  \city{Singapore}
  \country{Singapore}}
\email{echchang@ntu.edu.sg}

\renewcommand{\shortauthors}{Peipeng Yu et al.}

\begin{abstract}
  The proliferation of AIGC-driven face manipulation and deepfakes poses severe threats to media provenance, integrity, and copyright protection. Existing versatile watermarking systems typically rely on embedding explicit localization payloads, which introduces a fidelity--functionality trade-off: larger localization signals degrade visual quality and often reduce decoding robustness under strong generative edits. Moreover, these methods rarely support content recovery, limiting their forensic value when original evidence must be reconstructed. To address these challenges, we present VeriFi, a versatile watermarking framework that unifies copyright protection, pixel-level manipulation localization, and high-fidelity face content recovery. VeriFi makes three key contributions: (1) it embeds a compact semantic latent watermark that serves as a content-preserving prior, enabling faithful restoration even after severe manipulations; (2) it achieves fine-grained localization without dedicated payloads by correlating image features with decoded provenance signals; and (3) it introduces an AIGC attack simulator that combines latent-space mixing with seamless blending to enhance robustness against realistic deepfake pipelines. Extensive experiments on CelebA-HQ and FFHQ demonstrate that VeriFi consistently outperforms state-of-the-art baselines in watermark robustness, localization accuracy, and recovery quality, providing a practical and verifiable defense for deepfake forensics.
\end{abstract}



\maketitle

\section{Introduction}
\label{sec:intro}

Recent advances in artificial intelligence-generated content (AIGC) have enabled realistic face synthesis and accessible deepfakes, but they also raise serious concerns about misinformation, copyright infringement, and media authenticity~\cite{wang2025}. As generative models continue to improve, passive detectors struggle to generalize to sophisticated and unseen manipulations~\cite{tan2024frequency,tan2024rethinking,liu2025mun}, underscoring the need for proactive and verifiable defenses that remain robust under AIGC attacks.

\begin{figure}[tb]
\centering
\includegraphics[width=\linewidth]{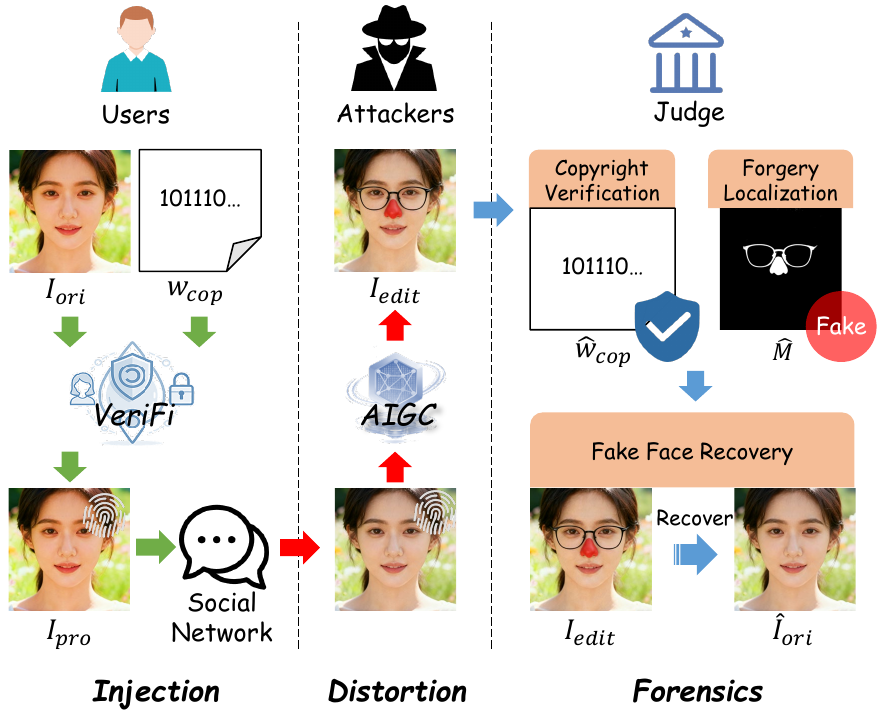}
\caption{Overview of our VeriFi framework for proactive deepfake defense. Our unified architecture embeds a robust copyright watermark and a compact semantic recovery prior into the original image $I_\mathrm{ori}$. For a potentially forged image $I_\mathrm{edit}$, VeriFi enables three integrated forensic capabilities: (1) provenance verification via robust copyright extraction; (2) pixel-level forgery localization, and (3) high-fidelity face recovery using the decoded semantic prior to reconstruct the original facial content $\hat{I}_\mathrm{ori}$.}
\label{fig:intro}
\end{figure}

Digital watermarking provides a proactive mechanism for establishing media provenance and safeguarding integrity~\cite{yang2024gaussian,zhao2024invisible}. Robust watermarks are designed to remain decodable under benign transformations and malicious edits (e.g., deepfake generation and splicing)~\cite{wang2021faketagger,wu2023sepmark,wang2024lampmark}, whereas fragile watermarks intentionally react to modifications, enabling sensitive tamper detection~\cite{neekhara2024facesigns,araghi2024disinformation,wang2025fractalforensics}. Building upon these paradigms, recent versatile frameworks (e.g., EditGuard~\cite{zhang2024editguard}, OmniGuard~\cite{zhang2025omniguard}, StableGuard~\cite{StableGuard}, and WAM~\cite{sander2025watermark}) seek to unify copyright protection with manipulation localization.
However, EditGuard and OmniGuard rely on high-capacity, dual-branch localization payloads, which increases embedding distortion and can weaken copyright robustness. In contrast, StableGuard and WAM largely depend on passive detectors, which often exhibit limited localization accuracy under unseen AIGC edits. Most importantly, these  methods remain irreversible, lacking a mechanism for high-fidelity content recovery for evidence preservation. 
How to escape the fidelity--functionality trade-off between copyright protection and tamper localization, while enabling high-fidelity content recovery, remains an open challenge.

To address these limitations, we propose \textbf{VeriFi}, a unified proactive framework that simultaneously supports (i) robust copyright protection, (ii) pixel-level tamper localization, and (iii) high-fidelity facial content recovery.
Unlike prior versatile methods that compromise visual quality and utility by using separate localization payloads, VeriFi introduces a novel watermark-conditioned localization network. By analyzing spatial inconsistencies directly within the decoded copyright signal, VeriFi eliminates the need for redundant localization watermarks.  Moreover, it incorporates a compact semantic recovery prior, enabling the accurate reconstruction of original facial features even after severe AIGC-based manipulation.
As illustrated in Figure~\ref{fig:intro}, given an original image $I_\mathrm{ori}$, our unified watermark encoder embeds a copyright code $w_\mathrm{cop}$ and a compact recovery representation ${z}_{face}$ to produce a protected image $I_{pro}$ for public distribution. In the event of a forged image $I_\mathrm{edit}$, VeriFi executes three core forensic operations: (1) \textbf{Provenance Verification:} it extracts the embedded copyright code $\hat{w}_\mathrm{cop}$ to verify the image source and ownership, maintaining robustness against common AIGC edits; (2) \textbf{Watermark-guided Localization:} It leverages the decoded $\hat{w}_\mathrm{cop}$ and $\hat{z}_{face}$ as spatial priors, mapping pixel-level forgeries by identifying deviation in the signal distribution; and (3) \textbf{High-fidelity Reconstruction:} It uses the decoded semantic recovery watermark $\hat{z}_{face}$ to synthesize a high-fidelity reconstruction $\hat{I}_\mathrm{ori}$, restoring the original facial content for evidence preservation. 

Our main contributions are summarized as follows:

\begin{itemize}
    \item We present \textbf{VeriFi}, a tamper-resilient versatile watermarking architecture that unifies robust copyright protection, pixel-level localization, and high-fidelity facial recovery. 

    \item We design a watermark-guided localization network that repurposes the decoded copyright signal as a spatial prior. This enables accurate tamper mapping without embedding localization-specific payloads, effectively resolving the inherent fidelity--functionality trade-off in prior works. 

    \item We introduce a watermark-guided recovery network that leverages a compact semantic watermark as a prior. This provides a strong structural guide for faithful restoration of the original face, even after severe AIGC manipulations.

    \item Extensive experiments on CelebA-HQ and FFHQ demonstrate that VeriFi outperforms state-of-the-art baselines in watermark robustness, localization precision, and reconstruction quality.
\end{itemize}

\section{Related Works}
\label{sec:related}

\subsection{Forgery Localization and Recovery Methods}

\begin{table}[tb]\large
\caption{Functional comparison of representative proactive watermarking and forensic methods. C: copyright protection; L: tamper localization; R: content recovery.}
\label{tab:compare}
\centering
\begin{adjustbox}{width=\linewidth}
\setlength{\tabcolsep}{2mm}
\renewcommand{\arraystretch}{1.05}
\begin{tabular}{lccccc}
\toprule
Method & Venue &Category & C & L  & R  \\
\midrule
SepMark \cite{wu2023sepmark}  & MM'23   & Face-aware robust watermarking  & \checkmark & -- & -- \\
LampMark \cite{wang2024lampmark} & MM'24   & Face-aware robust watermarking  & \checkmark & -- & -- \\
FakeTagger \cite{wang2021faketagger} & MM'21   & Face-aware robust watermarking  & \checkmark & -- & -- \\
AdaIFL \cite{li2024adaifl}       & ECCV'24  & Passive localization            & --         & \checkmark & -- \\
HiFi-Net \cite{guo2023hierarchical} & CVPR'23  & Passive localization            & -- & \checkmark & -- \\
Imuge+ \cite{ying2023learning}       & TPAMI'23  & Proactive localization and recovery   & -- & \checkmark & \checkmark \\
DFREC \cite{yu2025dfrecdeepfakeidentityrecovery} & TIFS'26     & Passive localization and recovery    & -- & \checkmark & \checkmark \\
EditGuard \cite{zhang2024editguard} & CVPR'24   & Versatile watermarking      & \checkmark & \checkmark & -- \\
OmniGuard \cite{zhang2025omniguard}    & CVPR'25   & Versatile watermarking      & \checkmark & \checkmark & -- \\
WAM \cite{sander2025watermark}          & ICLR'25   & Versatile watermarking      & \checkmark & \checkmark & -- \\
StableGuard \cite{StableGuard}  & NeurIPS'26   & Versatile watermarking      & \checkmark & \checkmark & -- \\
\midrule
VeriFi (ours)                    & --  & Versatile watermarking &    \checkmark & \checkmark & \checkmark \\
\bottomrule
\end{tabular}
\end{adjustbox}
\end{table}

Image forgery localization seeks to identify tampered regions at the pixel level across diverse manipulation types~\cite{zhou2023pre,zhang2024catmullrom,li2024adaifl,lou2025exploring,liu2025mun}. Representative passive methods such as MVSS-Net~\cite{dong2022mvss}, CAT-Net~\cite{kwon2021cat}, and HiFi-Net~\cite{guo2023hierarchical} leverage multi-branch architectures, frequency artifacts, and hierarchical attention to improve localization accuracy. 
However, these approaches often overfit to specific forgery distributions and do not support content recovery. While recent works like DFREC~\cite{yu2025dfrecdeepfakeidentityrecovery} extend passive localization to include recovery, they remain vulnerable to distribution shifts and complex AIGC manipulations. This highlights the need for a proactive strategy that encodes forensic priors into the content before manipulation occurs.
In this work, we propose a unified proactive framework that embeds robust watermarks prior to content distribution, which enables reliable tamper localization. Additionally, by simulating complex AIGC attacks, we further enhance watermark extraction robustness in  real-world applications.

\subsection{Proactive Watermarking Methods} 

\begin{figure*}[t]
    \centering
    \includegraphics[width=0.95\linewidth]{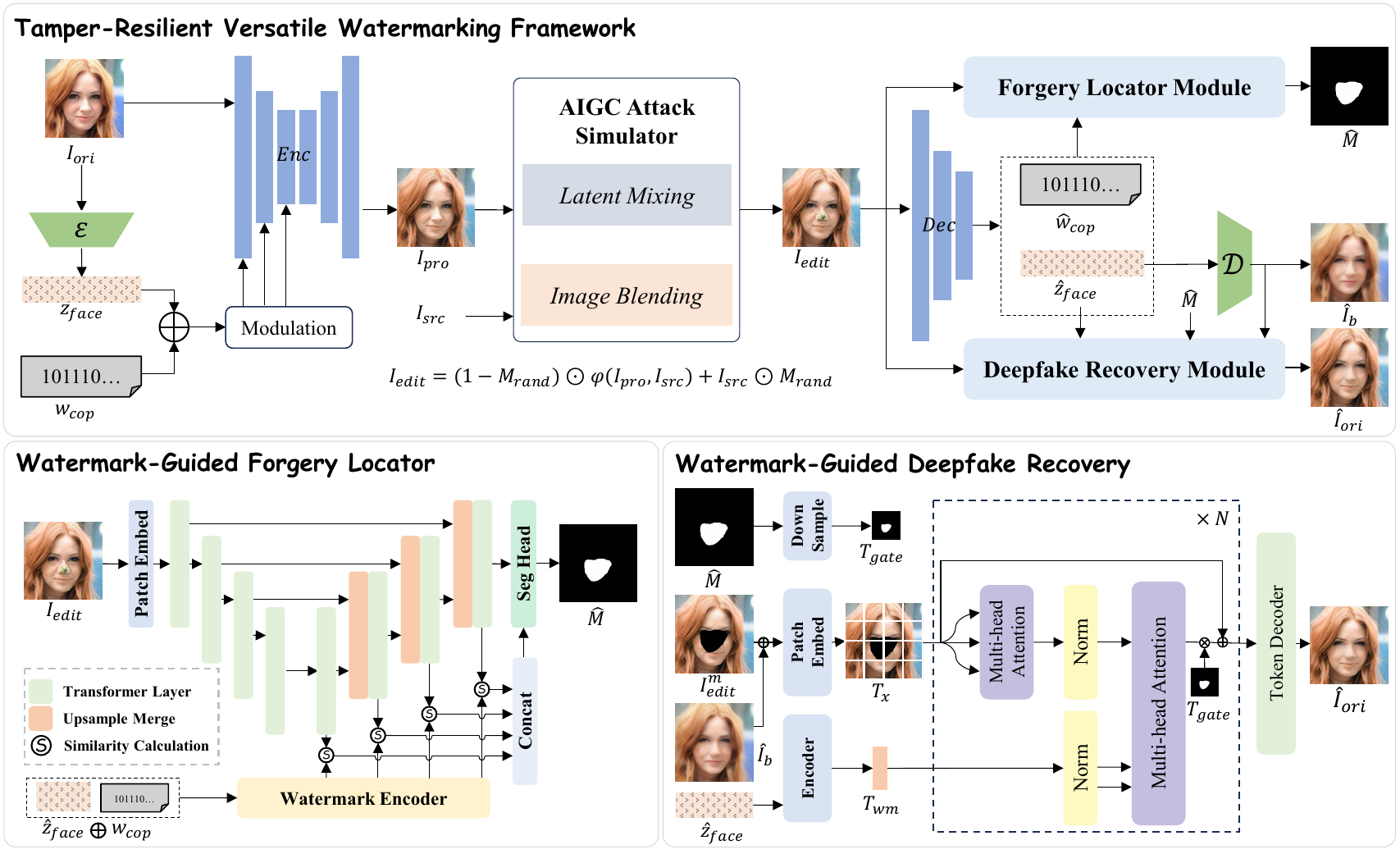}
    \caption{Overview of our \textbf{VeriFi}. (1) Unified Recovery-Copyright Watermark Embedder $Enc$ inserts an ownership code $w_{cop}$ and a compact facial signature $z_{face}$. (2) AIGC Attack Simulator performs Latent Mixing and Poisson blending to mimic realistic deepfake attacks during training. (3) 
    Watermark Extractor $Dec$ recovers $\hat{w}_{cop}$ and $\hat z_{face}$, reconstructs a coarse content proxy $\hat I_b$, and uses the decoded watermark signals to guide the Forgery Locator in predicting the manipulation map $\hat M$. 
    (4) Watermark-Guided Content Recovery Network employs a dual-stream Transformer with spatially gated cross-attention to fuse the edited image, $\hat I_b$, and $\hat M$ for selective content restoration.}
    \label{fig:framework}
\end{figure*}

Proactive watermarking methods embed imperceptible signals into images prior to distribution, enabling reliable provenance tracing and post-hoc forgery detection~\cite{wang2021faketagger,wu2023sepmark,wang2024lampmark,recovermark}. As summarized in Table~\ref{tab:compare}, existing approaches exhibit distinct design paradigms and technical focuses. Early deep watermarking methods, such as HiDDeN~\cite{zhu2018hidden} and StegaStamp~\cite{tancik2020stegastamp}, primarily target copyright protection, without supporting tamper localization or content recovery. While face-aware methods, including SepMark~\cite{wu2023sepmark} and LampMark~\cite{wang2024lampmark}, enhance robustness against facial manipulations, they remain limited to copyright verification. Conversely Imuge+~\cite{ying2023learning} extends proactive watermarking to tamper detection and content recovery via specialized perturbations but fails to address copyright authentication. Recent versatile frameworks, such as EditGuard~\cite{zhang2024editguard} and OmniGuard~\cite{zhang2025omniguard}, attempt to unify copyright protection and tamper localization. However, these methods rely on high-capacity localization payloads that can degrade visual fidelity and compromise robustness. While StableGuard~\cite{StableGuard} and WAM~\cite{sander2025watermark}  utilize passive locators to mitigate visual artifacts, they often exhibit limited generalization to unseen AIGC edits. RecoverMark \cite{recovermark} enables facial region restoration but lacks the ability of partial content recovery. Overall, existing solutions are constrained by inherent trade-offs among visual fidelity, adversarial robustness, and forensic functionality. 
VeriFi resolves these conflicts by introducing a watermark-conditioned architecture. By repurposing the copyright signal for localization and utilizing a compact semantic prior for reconstruction, we achieve comprehensive forensic capabilities without the payload-induced degradation characteristics of prior versatile frameworks.

\section{Proposed Method}
\subsection{Proposed Insights}
Existing deepfake watermarking methods face two core limitations: (1) a trade-off between copyright protection and precise tamper localization, and (2) a lack of unified frameworks for both authentication and content recovery. Our approach addresses these gaps through two key insights:

\noindent\textbf{Insight 1: Watermark degradation as a localization prior.} Content manipulations, such as face swapping, inevitably corrupt embedded watermarks within altered regions. This creates a spatial inconsistency: tampered areas exhibit damaged signals, while untampered areas remain intact. By detecting where the watermark fails to decode, we can precisely localize the manipulation.

\noindent\textbf{Insight 2: Semantic watermarking for high-fidelity recovery.}
Embedding a full original image (or its pixel-level representation) as a watermark requires excessive capacity, introducing visual distortion and poor robustness. 
Instead, we leverage Variational Autoencoder (VAE)-based compression to represent facial content as compact, low-dimensional latent codes. By embedding these semantic latent features, we enable high-fidelity  reconstruction while preserving watermark imperceptibility and  resilience.

\subsection{Overall Architecture}

Building on these insights, we introduce VeriFi, a unified, tamper-resilient framework for proactive deepfake defense that jointly supports copyright protection, pixel-level localization, and high-fidelity face recovery.
As shown in Figure~\ref{fig:framework}, VeriFi consists of four tightly coupled modules. At its core, the Unified Recovery-Copyright Watermark Embedder and Extractor (Sec.~\ref{sec:embedder_extractor}) jointly encode and decode two signals: a discrete $n$-bit ownership code $w_\mathrm{cop}$ for copyright authentication and a compact facial latent $z_\mathrm{face}$ derived from a pretrained VAE, which serves as a semantic recovery watermark. The embedder \textit{Enc} injects these signals into the image with minimal visual distortion and strong perturbation resilience, while the extractor \textit{Dec} is trained to reconstruct both signals even from manipulated inputs.
For tamper localization, the Watermark-Guided Forgery Localization Network (Sec.~\ref{sec:localization}) fuses image semantics with decoded watermark cues to predict a dense manipulation map $\hat{\mathbf{M}}$. Concurrently, the Watermark-Guided Content Recovery Network (Sec.~\ref{subsec:wm_recovery}) uses the recovered facial latent $\hat{z}_\mathrm{face}$ and $\hat{\mathbf{M}}$ to guide a dual-stream Transformer reconstructor $R$. Through cross-attention and adaptive modulation, $R$ restores tampered regions while preserving authentic content. During training, an AIGC Attack Simulator (Sec.~\ref{sec:aigc_simulator}) combines latent mixing, editing, and image blending to emulate realistic deepfake perturbations, improving robustness and generalization to diverse manipulations.

\subsection{Unified Recovery-Copyright Watermarking}
\label{sec:embedder_extractor}
To achieve robust copyright protection and high-fidelity content recovery, we design a Unified Recovery-Copyright Watermarking (URW) scheme. This module jointly embeds two complementary signals into the image: (i) a binary ownership code for copyright authentication, and (ii) a compact facial representation extracted by a pretrained Variational Autoencoder (VAE). This design enables visually imperceptible embedding while preserving robustness to common manipulations.

\noindent\textbf{Watermark construction:} Given an input image $I$, we first extract its VAE latent
$
z = E(I),
$
where $E(\cdot)$ denotes the pretrained VAE encoder. We then flatten $z$ into a 1D feature vector
$
z_{\mathrm{face}} = \mathrm{Flatten}(z) \in \mathbb{R}^{D},
$
which serves as the semantic facial watermark. In parallel, we define a binary ownership code
$
w_{\mathrm{cop}} \in \{0,1\}^{L}.
$
The two signals are concatenated to form the composite watermark
\begin{equation}
W = \big[\,z_{\mathrm{face}};\; w_{\mathrm{cop}}\,\big] \in \mathbb{R}^{D+L}.
\end{equation}

\noindent\textbf{Embedding:} Following \cite{bui2025trustmark}, we use an image-domain Watermark Embedder \textit{Enc} to inject $W$ into the cover image $I$, producing the watermarked image
\begin{equation}
I_\mathrm{wm} = Enc\big(I,\,W\big).
\end{equation}
Here, \textit{Enc} is implemented as a lightweight U-Net/residual network. The composite watermark $W$ is projected to encoder-compatible features and fused with image-level features to generate $I_\mathrm{wm}$ while preserving visual fidelity and robustness against common image degradations.

\noindent\textbf{Extraction:} Given a possibly manipulated image $I_{edit}$, a residual-based extractor \textit{Dec} recovers the embedded watermark:
\begin{equation}
\hat{W} = Dec(I_\mathrm{edit}) \in \mathbb{R}^{D+L}.
\end{equation}
We partition $\hat{W}$ into the first $D$ entries and the last $L$ entries, corresponding to the recovered facial latent $\hat{z}_{\mathrm{face}}$ and the recovered ownership code $\hat{w}_{\mathrm{cop}}$, respectively. $\hat{z}_{\mathrm{face}}$ is reshaped back to the original VAE latent tensor and decoded by a pretrained VAE decoder to obtain a coarse facial reconstruction $\hat{I}_{\mathrm{b}}$, while $\hat{w}_{\mathrm{cop}}$ undergoes a sigmoid transformation to produce bitwise probabilities for copyright verification. These recovered signals serve as priors for subsequent provenance authentication, tamper localization, and content recovery.

\subsection{Watermark-Guided Forgery Localization}
\label{sec:localization}
Tampering-induced degradation of embedded watermarks provides a strong spatial prior for manipulation detection. We propose a watermark-guided forgery localization network that explicitly aligns visual features with watermark-derived signals, enabling robust and precise identification of manipulated regions.
Our network adopts a dual-branch Swin-UNet architecture. The image branch extracts multi-scale semantic features from the potentially manipulated input, while the watermark branch encodes spatially aligned features decoded from the extracted watermarks ($\hat{z}_{\mathrm{face}}$ and $\hat{w}_{\mathrm{cop}}$). At each of $S$ hierarchical scales, both branches produce feature maps, which are projected to a common channel dimension $C_s$ using $1\times1$ convolutions to facilitate direct comparison.

To quantify the consistency between the two streams, we compute a per-pixel cosine similarity at each spatial location $(x, y)$, yielding a scale-specific similarity map $S_s$:
\begin{equation}
S_s(x,y)
=
\frac{\langle \mathcal{F}_{\mathrm{img}}^{(s)}(x,y), \, \mathcal{F}_{\mathrm{wm}}^{(s)}(x,y)\rangle}
{\|\mathcal{F}_{\mathrm{img}}^{(s)}(x,y)\|_2 \cdot \|\mathcal{F}_{\mathrm{wm}}^{(s)}(x,y)\|_2 },
\end{equation}
where $\mathcal{F}_{\mathrm{img}}^{(s)}\in\mathbb{R}^{C_s\times H_s\times W_s}$ and $\mathcal{F}_{\mathrm{wm}}^{(s)}\in\mathbb{R}^{C_s\times H_s\times W_s}$ denote the image and watermark feature maps at scale $s$, and $\langle\cdot,\cdot\rangle$ is the channel-wise inner product.

Subsequently, the similarity maps $\{S_1,\dots,S_S\}$ are upsampled to the input resolution and aggregated by a lightweight fusion head:
\begin{equation}
\mathcal{F}_{\mathrm{sim}} = g\!\left(\mathrm{Concat}\!\big(\mathrm{Up}(S_1),\dots,\mathrm{Up}(S_S)\big)\right),
\end{equation}
where $\mathrm{Up}(\cdot)$ denotes spatial upsampling and $g(\cdot)$ is a stack of convolutional layers with normalization that produces a dense evidence map.

Finally, we fuse the model-driven decoder evidence and the watermark-aligned evidence via a learnable gated mechanism. The final manipulation probability map $\hat{\mathbf{M}}$ is computed as:
\begin{equation}
\hat{\mathbf{M}} = \sigma\big(\mathcal{F}_{\mathrm{dec}} + \alpha\,\mathcal{F}_{\mathrm{sim}}\big),
\end{equation}
where $\mathcal{F}_{\mathrm{dec}}$ is the semantic decoder output, $\sigma(\cdot)$ is the segmentation head, and $\alpha$ is a trainable scalar constrained to $[0,1]$.

\subsection{Watermark-Guided Content Recovery}
\label{subsec:wm_recovery}
The URW Embedder encodes a compact facial latent that remains decodable after heavy manipulations, serving as a semantic prior for recovery. A coarse face proxy is reconstructed from $\hat{z}_{face}$ via a pretrained VAE decoder. However, direct reconstruction from this proxy is typically over-smoothed and lacks fine-grained detail. To address this, we introduce a watermark-guided recovery network that adaptively fuses spatial features from the manipulated image with semantic priors from the watermark, enabling high-fidelity restoration of the original face.

As illustrated in Figure~\ref{fig:framework}, the recovery network receives four inputs: (1) the manipulated image $I^{m}_\mathrm{edit}$ (masked according to the predicted $\hat{M}$), (2) a coarse content proxy $\hat{I}_b = D_{\text{VAE}}(\hat{z}_\mathrm{face})$ reconstructed from the decoded semantic watermark, (3) the decoded facial watermark $\hat{z}_\mathrm{face}$, and (4) the predicted manipulation mask $\hat{M}$. The proxy $\hat{I}_b$ provides structural and content-consistent cues that complement the corrupted observation $I^{m}_\mathrm{edit}$. We concatenate $\hat{I}_b$ with $I^{m}_\mathrm{edit}$ and apply a patch embedding to generate image tokens $T_x$, which are then processed by a stack of $N$ Transformer blocks for feature fusion and recovery.
In parallel, $\hat{z}_\mathrm{face}$ is mapped into a set of semantic tokens $T_\mathrm{wm}$ via a dedicated watermark encoder. To ensure that semantic priors are injected exclusively into manipulated regions, the predicted mask $\hat{M}$ is downsampled to the token resolution and used as a spatial gating mechanism during cross-attention. Formally, the gated cross-attention at each Transformer block is defined as:
\begin{equation}
T'_x = T_x + \mathcal{G}(\hat{M}) \odot \sum_{h} \text{Attn}\big(\text{LN}(T_x),\, \text{LN}(T_\mathrm{wm}),\, \text{LN}(T_\mathrm{wm})\big),
\end{equation}
where $\mathcal{G}(\cdot)$ is a spatial gating operator, $\text{Attn}(\cdot)$ denotes the cross-attention mechanism for head $h$, and $\text{LN}(\cdot)$ denotes layer normalization. This mechanism restricts watermark-guided feature fusion to predicted tampered regions, preventing semantic leakage into authentic areas. After $N$ Transformer layers, a lightweight decoder reconstructs the recovered image $\hat{I}_\mathrm{ori}$.

\subsection{AIGC Attack Simulator}
\label{sec:aigc_simulator}
Modern deepfake pipelines typically comprise latent-space synthesis followed by image-space blending. Existing watermarking methods mainly model post-blending perturbations while overlooking synthesis-induced feature distortions~\cite{zhang2025omniguard}. To address this limitation, we introduce an AIGC Attack Simulator that jointly models latent-space and image-space manipulations during training. This simulator exposes the watermarking network to more realistic and challenging attack scenarios. The simulator is used only during training and requires no prior knowledge of AIGC tools encountered at inference.

\begin{figure}[t]
    \centering
    \includegraphics[width=\linewidth]{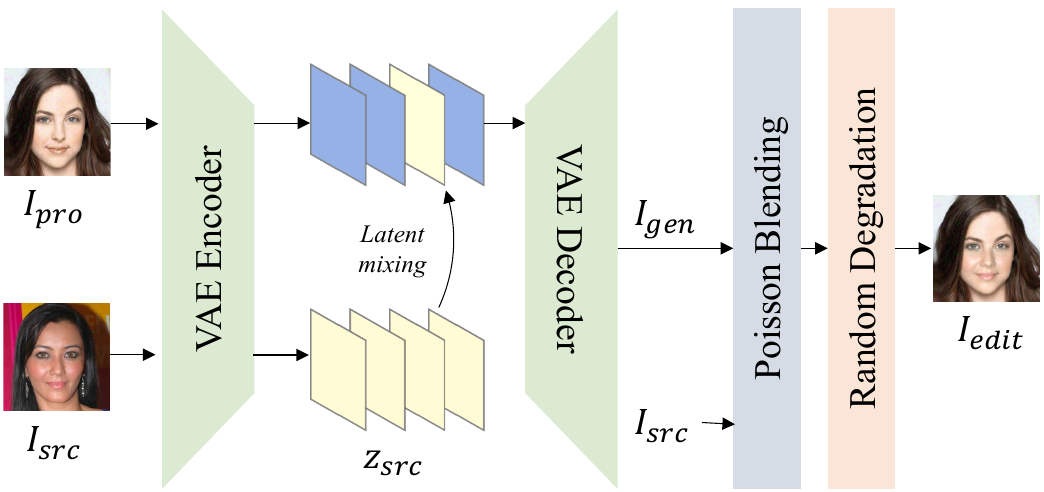}
    \caption{Overview of the proposed AIGC Attack Simulator. It emulates AIGC edits through latent feature grafting and seamless image blending to improve watermark robustness.}
    \label{fig:simulate}
\end{figure}

\noindent\textbf{Latent Mixing.} To emulate the feature blending process inherent in modern deepfake generators, we design a Latent Mixing Module, simulating the latent-space manipulations commonly performed by AIGC models. As depicted in Figure~\ref{fig:simulate}, given a watermarked image $\mathbf{I}_\mathrm{pro}$, we first select a source face $\mathbf{I}_\mathrm{src}$ with similar facial landmarks~\cite{shiohara2022detecting,yuunlocking}, and extract their latent representations $\mathbf{z}_\mathrm{pro}, \mathbf{z}_\mathrm{src} \in \mathbb{R}^{C \times H \times W}$ using a pretrained VAE Encoder. We then randomly sample a binary channel mask $\mathbf{M}_{c} \in \{0,1\}^{C}$, and construct the mixed latent code as
\begin{equation}
\mathbf{z}_{\mathrm{gen}} = \mathbf{M}_{c} \odot \mathbf{z}_\mathrm{src} + (1-\mathbf{M}_{c}) \odot \mathbf{z}_\mathrm{pro},
\end{equation}
where $\odot$ denotes element-wise multiplication broadcasted across spatial dimensions. The resulting $\mathbf{z}_{\mathrm{gen}}$ is decoded by the VAE Decoder to obtain the synthesized image $\mathbf{I}_{\mathrm{gen}}$. This process incorporates model-specific artifacts and external features, thereby more accurately simulating AIGC editing.

\noindent\textbf{Image Blending.} To simulate realistic face replacement, we integrate the synthesized face $\mathbf{I}{\mathrm{gen}}$ into the original background via Poisson blending~\cite{perez2003poisson}. Following LVLM-DFD~\cite{yuunlocking}, we generate random facial masks $\mathbf{M}$ from detected landmarks and solve the Poisson equation within the masked regions, yielding seamlessly blended images $\mathbf{I}{\mathrm{edit}}$. We further apply random degradations, such as JPEG compression and Gaussian noise, to improve forensic robustness.

\subsection{Training Objectives}

\paragraph{Watermark Embedding.}
To guarantee visual imperceptibility, we constrain both pixel-level distortion and perceptual discrepancy between the original image $\mathbf{I}_\mathrm{ori}$ and the watermarked image $\mathbf{I}_\mathrm{wm}$:
\begin{equation}
    \mathcal{L}_\text{embed} = \lVert \mathbf{I}_\mathrm{wm} - \mathbf{I}_\mathrm{ori} \rVert_2^2 + \sum_{\ell} \lVert \phi_\ell(\mathbf{I}_\mathrm{wm}) - \phi_\ell(\mathbf{I}_\mathrm{ori}) \rVert_1,
\end{equation}
where $\phi_\ell$ denotes features extracted from the $\ell$-th layer of a pretrained VGG-19 network.

\paragraph{Message Decoding.}
To ensure robust recovery of both the ownership code $w_\mathrm{cop}$ and the facial signature $z_\mathrm{face}$ from manipulated images, we jointly optimize:
\begin{equation}
    \mathcal{L}_\text{decode} = \mathrm{BCE}(\hat{w}_\mathrm{cop}, w_\mathrm{cop}) + \lVert \hat{z}_\mathrm{face} - z_\mathrm{face} \rVert_2^2,
\end{equation}
where $\hat{w}_\mathrm{cop}$ and $\hat{z}_\mathrm{face}$ are the decoded ownership code and facial signature, respectively.

\paragraph{Forgery Localization.}
Manipulation localization is supervised using the Dice Loss~\cite{7785132} as our localization objective $\mathcal{L}_\text{loc}$, which directly optimizes the overlap between the predicted manipulation map $\hat{\mathbf{M}}$ and the ground-truth mask $\mathbf{M}^*$. 
\paragraph{Guided Recovery.}
Restoration fidelity is enforced by constraining the reconstructed image $\hat{\mathbf{I}}_\mathrm{ori}$ to match the original $\mathbf{I}_\mathrm{ori}$ in both pixel and perceptual domains:
\begin{equation}
    \mathcal{L}_\text{rec} = \lVert \hat{\mathbf{I}}_\mathrm{ori} - \mathbf{I}_\mathrm{ori} \rVert_1 + \sum_{\ell} \lVert \phi_\ell(\hat{\mathbf{I}}_\mathrm{ori}) - \phi_\ell(\mathbf{I}_\mathrm{ori}) \rVert_1.
\end{equation}

\paragraph{Overall Objective.}
The complete training objective is formulated as:
\begin{equation}
    \mathcal{L}_\text{total} = \mathcal{L}_\text{embed} + \lambda_1 \mathcal{L}_\text{decode} + \lambda_2 \mathcal{L}_\text{loc} + \lambda_3 \mathcal{L}_\text{rec},
\end{equation}
where $\lambda_i$ are hyperparameters balancing the respective loss terms.

\section{Experiments}
\label{sec:experiments}

\subsection{Experimental Settings}
\label{sec:exp_settings}

\noindent\textbf{Datasets.} 
We evaluate our method on CelebA-HQ~\cite{karras2018progressive} and FFHQ~\cite{karras2019style}. CelebA-HQ contains 30,000 high-resolution celebrity images, for which we adopt the official train/validation/test split. FFHQ comprises 70,000 diverse, high-quality face images; we use 65,000/4,000/ 1,000 images for training/validation/testing. All images are resized to $256 \times 256$.

\noindent\textbf{Implementation Details.} We use Adam optimizer with a learning rate of $2 \times 10^{-4}$ and batch size 8. The loss weights are set as $\lambda_{1}=1.0$, $\lambda_{2}=1.0$, and $\lambda_{3}=2.0$.  All experiments are conducted on four NVIDIA RTX 3090 GPUs. We use SD Inpainting~\cite{rombach2022high}, HD-painter~\cite{manukyan2023hd}, E4S~\cite{liu2023fine}, InfoSwap~\cite{Gao_2021_CVPR}, MaskFaceGAN~\cite{10299582}, and SimSwap~\cite{chen2020simswap} as the AIGC attack models to evaluate performance under diverse manipulation scenarios. Additional implementation details are provided in the Supplementary Material.

\noindent\textbf{Baselines.} We compare our method against state-of-the-art deep watermarking,  tamper localization and image recovery methods, including: SepMark~\cite{wu2023sepmark}, TrustMark~\cite{bui2025trustmark}, EditGuard~\cite{zhang2024editguard}, Robust-Wide~\cite{hu2024robust}, OmniGuard~\cite{zhang2025omniguard}, StableGuard \cite{StableGuard}, WAM \cite{sander2025watermark}, CAT-Net~\cite{kwon2021cat}, PSCC-Net~\cite{liu2022pscc}, HiFi-Net~\cite{guo2023hierarchical}, MVSS-Net\cite{dong2022mvss}, NCL-IML~\cite{zhou2023pre}, Imuge+~\cite{ying2023learning}, and DFREC \cite{yu2025dfrecdeepfakeidentityrecovery}. All baselines are implemented using their official codebases and pretrained weights.

\noindent\textbf{Evaluation Metrics.} Imperceptibility is evaluated by Peak Signal-to-Noise Ratio (PSNR$\uparrow$) and Structural Similarity Index (SSIM$\uparrow$). Watermark robustness is quantified by Bit Accuracy$\uparrow$ (\%), the fraction of correctly decoded bits under each attack. Tamper localization is assessed using F1-score$\uparrow$, Area Under the Curve (AUC$\uparrow$), and mean Intersection-over-Union (mIoU$\uparrow$). Recovery quality is measured by $\emph{PSNR}_\emph{rec}$$\uparrow$/$\emph{SSIM}_\emph{rec}$$\uparrow$, and Fréchet Inception Distance (FID$\downarrow$).

\begin{table}[t]
\centering
\scriptsize
\caption{Tamper-localization results on CelebA-HQ (1,000 images). Metrics: F1 / AUC / mIoU (higher is better). Best and second-best per column are \textbf{bold} and \underline{underlined}, respectively.}
\label{tab:localization}
\renewcommand\arraystretch{1.1}
\setlength{\tabcolsep}{1.5pt}
\resizebox{\columnwidth}{!}{%
\begin{tabular}{@{}llccc|ccc|ccc@{}}
\toprule
Method & Venue & \multicolumn{3}{c}{SD Inpainting} & \multicolumn{3}{c}{HD-painter} & \multicolumn{3}{c}{Splicing} \\
\cmidrule(lr){3-5}\cmidrule(lr){6-8}\cmidrule(lr){9-11}
 &  & F1 & AUC & mIoU & F1 & AUC & mIoU & F1 & AUC & mIoU \\
\midrule
CAT-Net\cite{kwon2021cat}   & WACV'21 & 0.066 & 0.714 & 0.427 & 0.107 & 0.727 & 0.447 & 0.490 & 0.876 & 0.659 \\
MVSS-Net\cite{dong2022mvss}  & TPAMI'22 & 0.262 & 0.843 & 0.484 & 0.378 & 0.866 & 0.540 & 0.639 & 0.935 & 0.707 \\
PSCC-Net\cite{liu2022pscc}  & TCSVT'22 & 0.310 & 0.702 & 0.275 & 0.335 & 0.691 & 0.245 & 0.362 & 0.727 & 0.290 \\
HiFi-Net\cite{guo2023hierarchical}  & CVPR'23 & 0.115 & 0.742 & 0.430 & 0.396 & 0.815 & 0.567 & 0.169 & 0.757 & 0.458 \\
NCL-IML\cite{zhou2023pre}   & ICCV'23 & 0.085 & 0.518 & 0.423 & 0.087 & 0.516 & 0.421 & 0.040 & 0.503 & 0.406 \\
Imuge+\cite{ying2023learning}    & TPAMI'23 & 0.414 & 0.854 & 0.580 & 0.699 & 0.943 & 0.764 & 0.719 & 0.926 & 0.792 \\
EditGuard\cite{zhang2024editguard} & CVPR'24 & 0.862 & 0.894 & 0.864 & 0.849 & 0.869 & 0.818 & 0.867 & 0.895 & 0.869 \\
OmniGuard\cite{zhang2025omniguard} & CVPR'25 & \underline{0.869} & \textbf{0.983} & \underline{0.880} & \underline{0.914} & \textbf{0.995} & \underline{0.908} & 0.804 & 0.980 & 0.817 \\
WAM\cite{sander2025watermark}       & ICLR'25 & 0.495 & 0.787 & 0.492 & 0.174 & 0.503 & 0.315 & 0.782 & 0.921 & 0.707 \\
StableGuard\cite{StableGuard} & NeurIPS'26 & 0.686 & 0.482 & 0.548 & 0.707 & 0.384 & 0.570 & \textbf{0.997} & \textbf{1.000} & \textbf{0.994} \\

VeriFi (Ours)
           & Ours & \textbf{0.946} & \underline{0.975} & \textbf{0.941}
           & \textbf{0.975} & \underline{0.975} & \textbf{0.939}
           & \underline{0.989} & \underline{0.995} & \underline{0.985} \\
\bottomrule
\end{tabular}%
}
\end{table}

\subsection{Comparison with Localization Methods}
\label{sec:exp_localization}
We evaluate tamper-localization performance on 1,000 CelebA-HQ images under three representative AIGC-based forgery types: SD Inpainting, HD-painter and Face Splicing. Comparisons include recent passive detectors (CAT-Net~\cite{kwon2021cat}, PSCC-Net~\cite{liu2022pscc}, HiFi-Net~\cite{guo2023hierarchical}, NCL-IML~\cite{zhou2023pre}) and active watermarking and recovery-aware baselines (EditGuard~\cite{zhang2024editguard}, OmniGuard~\cite{zhang2025omniguard}, Imuge+~\cite{ying2023learning}, StableGuard~\cite{StableGuard}, WAM~\cite{sander2025watermark}). Table~\ref{tab:localization} reports F1, AUC and mIoU for each tampering scenario. 
Passive methods exhibit limited sensitivity to subtle generative alterations, while several watermarking baselines degrade noticeably under AIGC-induced perturbations. Notably, VeriFi achieves F1 scores of 0.946, 0.975, and 0.989 on SD Inpainting, HD-painter, and Splicing, respectively, overall outperforming prior methods.
Qualitative examples in the Supplementary Material corroborate these quantitative findings and illustrate clearer boundaries for VeriFi compared to prior work.
We also include additional experiments on FFHQ dataset in the Supplementary Material. Both quantitative and qualitative results consistently demonstrate VeriFi's superior tamper-localization capabilities across diverse datasets and manipulation types.

\begin{figure}[t]
    \centering
    
    \includegraphics[width=\linewidth]{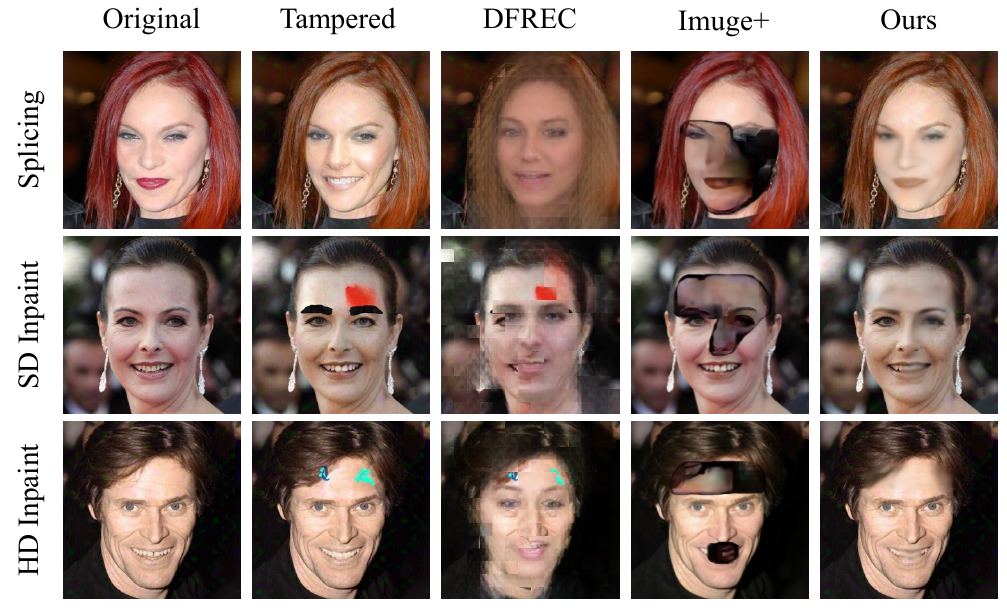}
    \caption{Qualitative comparison of face recovery under SD Inpainting/HD-painter and Splicing.}
    \label{fig:recovery_results}
\end{figure}

\begin{figure}[h]
    \centering
    \includegraphics[width=\linewidth]{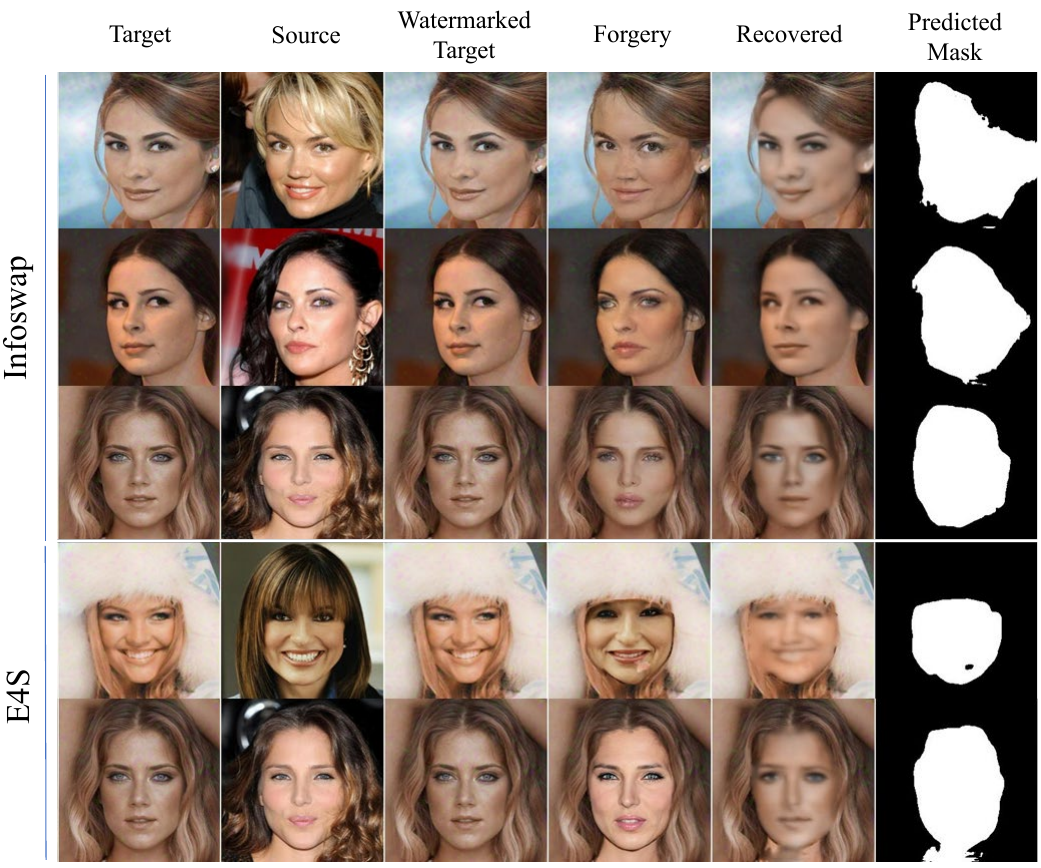}
    \caption{Face recovery visualization under InfoSwap/E4S  on CelebA-HQ.  }
    \label{fig:additional_recovery_celeba_infoswap}
\end{figure}

\begin{table*}[h]
\centering
\caption{Face recovery performance under six representative tampering types. Higher $\emph{PSNR}_\emph{rec}$/$\emph{SSIM}_\emph{rec}$ indicate better fidelity; lower FID indicates higher realism. Attack types are grouped by generation paradigm: \textbf{Diffusion-based} (SD Inpainting, HD-painter), \textbf{GAN-based} (E4S, InfoSwap, MaskFaceGAN), and \textbf{Traditional} (Splicing). Best results per column are \textbf{bold}.}
\label{tab:face_recovery_extended}
\scriptsize
\renewcommand\arraystretch{1.15}
\setlength{\tabcolsep}{0.8pt}
\resizebox{\linewidth}{!}{%
\begin{tabular}{l|c|ccc|ccc|ccc|ccc|ccc|ccc}
\toprule
\multirow{2}{*}{Method} & \multirow{2}{*}{Dataset} & \multicolumn{6}{c|}{\textbf{Diffusion-based}} & \multicolumn{9}{c|}{\textbf{GAN-based}} &   \multicolumn{3}{c}{\textbf{Traditional}}\\
\cmidrule(r){3-8}\cmidrule(l){9-20}
    & & \multicolumn{3}{c|}{SD Inpainting} & \multicolumn{3}{c|}{HD-painter} & \multicolumn{3}{c|}{E4S} & \multicolumn{3}{c|}{InfoSwap} & \multicolumn{3}{c|}{MaskFaceGAN} & \multicolumn{3}{c}{Splicing} \\
\cmidrule(r){3-5}\cmidrule(r){6-8}\cmidrule(r){9-11}\cmidrule(r){12-14}\cmidrule(r){15-17}\cmidrule(l){18-20}
    & & $\emph{PSNR}_\emph{rec}$ & $\emph{SSIM}_\emph{rec}$ & FID & $\emph{PSNR}_\emph{rec}$ & $\emph{SSIM}_\emph{rec}$ & FID & $\emph{PSNR}_\emph{rec}$ & $\emph{SSIM}_\emph{rec}$ & FID & $\emph{PSNR}_\emph{rec}$ & $\emph{SSIM}_\emph{rec}$ & FID & $\emph{PSNR}_\emph{rec}$ & $\emph{SSIM}_\emph{rec}$ & FID & $\emph{PSNR}_\emph{rec}$ & $\emph{SSIM}_\emph{rec}$ & FID \\
\midrule
Imuge+\cite{ying2023learning}           & CelebA-HQ & 21.49          & 0.741           & \textbf{29.60}  & 16.91          & 0.753            & 118.07         & 18.52          & 0.625           & 129.94         & 17.69          & 0.674           & 68.41        & 19.65          & 0.622          & 105.23         & 27.07          & 0.906          & 34.52          \\
DFREC\cite{yu2025dfrecdeepfakeidentityrecovery}           & CelebA-HQ & 21.14           & 0.666          & 71.68          & 18.36          & 0.691            & 107.15         & 17.74          & 0.608          & 38.57          & 21.11          & 0.628          & 38.51          & 18.63          & 0.624         & 91.03         & 19.64          & 0.661         & 59.95          \\
VeriFi   (Ours) & CelebA-HQ & \textbf{31.05}  & \textbf{0.894}  & 39.25          & \textbf{28.34} & \textbf{0.892}   & \textbf{45.68} & \textbf{25.12} & \textbf{0.839}  & \textbf{26.71} & \textbf{27.33} & \textbf{0.866} & \textbf{18.82} & \textbf{25.73} & \textbf{0.845} & \textbf{41.18} & \textbf{31.06} & \textbf{0.921} & \textbf{21.51} \\
\midrule
Imuge+\cite{ying2023learning}           & FFHQ      & 16.46        & 0.745          & 94.05        & 19.54          & 0.697            & 47.13          & 11.28          & 0.542           & 121.56         & 15.32        & 0.592          & 78.10       & 18.34          & 0.619          & 63.57          & 29.52       & 0.891        & 31.48          \\
DFREC\cite{yu2025dfrecdeepfakeidentityrecovery}           & FFHQ      & 20.90           & 0.656          & 73.03          & 10.47          & 0.342           & 199.65         & 16.64          & 0.577          & 50.04          & 20.33          & 0.598          & 49.32          & 17.19          & 0.575         & 127.08         & 17.26          & 0.606        & 71.49          \\
VeriFi   (Ours) & FFHQ      & \textbf{30.61} & \textbf{0.905} & \textbf{20.21} & \textbf{26.84} & \textbf{0.853} & \textbf{29.83} & \textbf{21.1}  & \textbf{0.788} & \textbf{45.36} & \textbf{24.25} & \textbf{0.766} & \textbf{38.84} & \textbf{25.19} & \textbf{0.866} & \textbf{38.20}  & \textbf{31.60}  & \textbf{0.937} & \textbf{23.56} \\
\bottomrule
\end{tabular}%
}
\end{table*}

\begin{table*}[h]
\centering
\scriptsize
\caption{Quantitative comparison on CelebA-HQ. 
(PSNR in dB/SSIM) and watermark extraction accuracy (\%) under representative AIGC manipulations and common degradations.
For each column, the best and second-best results are highlighted in \textbf{bold} and \underline{underlined}, respectively. $W$ and $R$ denote auxiliary localization and recovery watermarks.}
\label{tab:watermarking}
\renewcommand\arraystretch{1.1}
\begin{adjustbox}{width=\linewidth}
\setlength{\tabcolsep}{1mm}
\begin{tabular}{@{}lcccc|ccc|cc|ccc|c@{}}
\toprule
\multirow{2}{*}{Method} & \multirow{2}{*}{Venue}& \multirow{2}{*}{Capability} & \multirow{2}{*}{PSNR} & \multirow{2}{*}{SSIM} & \multicolumn{3}{c|}{GAN-based (\%)} & \multicolumn{2}{c|}{Diffusion-based (\%)} & \multicolumn{3}{c|}{Common degradations (\%)} & \multirow{2}{*}{Avg. (\%)} \\
 &  & & &  & InfoSwap & E4S & MaskFaceGAN & HD-painter & SD Inpainting & JPEG & ColorJitter & Gaussian Noise &  \\
\midrule

SepMark~\cite{wu2023sepmark} & MM'23& 48bits & 38.67 & 0.928 & 92.01 & 88.49 & \underline{95.02} & \underline{98.83} & 98.35 & \underline{99.99} & \underline{99.98} & \textbf{100.00} & 96.58 \\
LampMark~\cite{wang2024lampmark} & MM'24 &64bits & 44.05 & \textbf{0.971} & 89.21 & 74.80 & 65.61 & 83.73 & 89.79 & 84.13 & 84.50 & 84.88 & 82.08 \\
Robust-Wide~\cite{hu2024robust} & ECCV'24 &64bits & 43.74 & 0.952 & 86.31 & 86.48 & 86.90 & 97.20 & 96.09 & 96.45 & 97.15 & 97.43 & 93.00 \\
EditGuard~\cite{zhang2024editguard} & CVPR'24 &64bits+$W$ & 32.24 & 0.758 & 48.92 & 77.34 & 78.56 & 82.07 & 91.39 & 79.55 & 95.07 & 93.19 & 80.76 \\
OmniGuard~\cite{zhang2025omniguard} & CVPR'25 &100bits+$W$ & 42.08 & 0.947 & 72.32 & 73.66 & 79.80 & 76.77 & 84.33 & 92.33 & 92.13 & 92.29 & 82.95 \\
TrustMark~\cite{bui2025trustmark} & ICCV'25 &100bits & 47.94 & 0.955 & 65.03 & 68.76 & 53.01 & 90.22 & 87.21 & 96.31 & 98.96 & 99.81 & 82.41 \\
WAM~\cite{sander2025watermark} & ICLR'25 &48bits& 43.47 & 0.950 & \underline{99.64} & \textbf{99.87} & 92.97 & 77.60 & 93.34 & \textbf{100.00} & 59.64 & \underline{99.99} & 90.38 \\
StableGuard~\cite{StableGuard} & NeurIPS'26 &48bits & 31.81 & 0.848 & 50.08 & 87.93 & 89.16 & 88.13 & 95.58 & 59.78 & 99.93 & 63.61 & 79.27 \\
\midrule
VeriFi   (Ours)                                           & -                      & 48bits                      & \textbf{50.49}       & \underline{0.959}    & \textbf{99.85} & \underline{96.13} & \textbf{96.14} & \textbf{99.34} & \textbf{98.81} & 99.93 & \textbf{100.00} & \textbf{100.00} & \textbf{98.78} \\
VeriFi   (Ours)                                           & -                      & 100bits                     & \underline{48.14}    & 0.953                 & 98.92 & 93.89 & 94.87 & 98.76 & \underline{98.64} & 99.91 & \textbf{100.00} & \textbf{100.00} & \underline{98.12} \\
VeriFi   (Ours)                                           & -                      & 100bits + $R$               & 42.53                 & 0.948                 & 99.60 & 90.51 & \underline{95.62} & 98.72 & 97.29 & 99.82 & \textbf{100.00} & \textbf{100.00} & 97.70 \\                
\bottomrule
\end{tabular}
\end{adjustbox}
\end{table*}

\subsection{Comparison with Face Recovery Methods}
\label{sec:exp_recovery}

We comprehensively evaluate face recovery performance by benchmarking VeriFi against state-of-the-art methods, including Imuge+ \cite{ying2023learning} and DFREC \cite{yu2025dfrecdeepfakeidentityrecovery}, on the CelebA-HQ dataset. As summarized in Table~\ref{tab:face_recovery_extended}, VeriFi consistently achieves the best overall results, with substantial margins over prior works. VeriFi attains 31.05 dB PSNR and 0.894 SSIM on CelebA-HQ under SD Inpainting, outperforming Imuge+ and DFREC. Similar trends are observed for HD-painter and splicing, as well as on FFHQ, demonstrating the robustness of our approach.
Qualitative results in Figures~\ref{fig:recovery_results} and \ref{fig:additional_recovery_celeba_infoswap} further highlight that VeriFi can faithfully reconstruct fine-grained facial details, even under challenging and large-area tampering. In contrast, existing methods often produce visible artifacts or fail to recover occluded regions. Notably, VeriFi maintains high visual quality and low FID, indicating that the recovered faces are not only visually plausible but also semantically consistent with the original content. More recovery results under diverse AIGC manipulations and datasets are provided in the Supplementary Material, further substantiating the effectiveness of VeriFi in real-world forgery scenarios.

\subsection{Comparison with Deep Watermarking}
\label{sec:exp_watermarking}

We compare VeriFi against eight state-of-the-art deep watermarking methods~\cite{bui2025trustmark,wu2023sepmark,hu2024robust,zhang2024editguard,zhang2025omniguard,StableGuard,sander2025watermark,wang2024lampmark}. Table~\ref{tab:watermarking} reports both watermarked-image fidelity and message extraction accuracy under representative AIGC manipulations and common degradations on CelebA-HQ. We evaluate multiple VeriFi variants with different embedding capacities (48 bits and 100 bits) and functionalities (with or without recovery watermark) to assess the impact of these factors on performance. The results demonstrate that increasing the embedding capacity or adding the recovery watermark does not significantly affect extraction accuracy, indicating that our design achieves strong extensibility while maintaining high robustness. When comparing across methods, VeriFi achieves the best overall performance, with an average extraction accuracy of 98.78\% for the 48-bit variant and 98.12\% for the 100-bit variant, substantially outperforming prior methods.
Although some baselines achieve slightly stronger fidelity metrics in isolation, their decoding robustness degrades markedly under challenging manipulations. 
Overall, these results indicate that VeriFi offers a more favorable operating point than prior methods, preserving high perceptual fidelity while sustaining reliable message recovery under both generative edits and conventional degradations. Additional qualitative comparisons, cross-dataset generalization results on FFHQ and ImageNet, different resolution settings, and extended robustness evaluations are provided in the Supplementary Material.

\subsection{Analysis and Ablation Studies}
\noindent\textbf{Ablation on AIGC Attack Simulation.}
We quantify the individual and combined effects of the simulator components, namely noise augmentation, image blending, and latent  mixing, by training four variants: (i) without augmentation, (ii) with noise augmentation, (iii) with image blending, and (iv) with both blending and latent  mixing, which constitutes the full model referred to as Ours. Table~\ref{tab:aigc_ablation_attacks} reports bit extraction accuracy (\%) on 1,000 CelebA-HQ images under four representative attacks. The results show that both image blending and latent  mixing are required to realistically simulate AIGC attacks and to obtain robust watermark extraction.

\begin{table}[t]\huge
\centering
\setlength{\tabcolsep}{1mm}
\caption{Ablation of AIGC attack-simulator components. Bit extraction accuracy (\%) under four representative attacks on 1,000 CelebA-HQ images.}
\label{tab:aigc_ablation_attacks}
\renewcommand\arraystretch{1.2}
\begin{adjustbox}{max width=\linewidth}
\begin{tabular}{@{}lccccc@{}}
\toprule
Method & E4S & SimSwap & SD Inpainting & InfoSwap & Avg. \\
\midrule
Without augmentation & 50.97 & 53.04 & 63.69 & 54.11 & 55.45 \\
With noise augmentation & 57.77 & 66.63 & 79.27 & 73.26 & 69.23 \\
With blending & 85.47 & 76.34 & \textbf{87.95} & 84.16 & 83.48 \\
Blending + latent mixing (Ours) & \textbf{94.25} & \textbf{77.52} & 86.45 & \textbf{87.90} & \textbf{86.53} \\
\bottomrule
\end{tabular}
\end{adjustbox}
\end{table}

\noindent\textbf{Ablation on Similarity Computation for Localization.}
We evaluate the effect of augmenting the localization head with a similarity branch in Table~\ref{tab:loc_similarity_metric}. Replacing segmentation-only supervision with an additional similarity branch improves F1 from 0.912 to 0.946, AUC from 0.966 to 0.975, and mIoU from 0.842 to 0.941, demonstrating that watermark-guided similarity cues substantially enhance localization accuracy. We further compare two pairing strategies during training: forming pairs from original (genuine) images and their watermarked counterparts, and forming pairs by concatenating unrelated images with watermarked images. Training with a mixture of genuine and synthetic composite pairs yields the largest improvement in localization performance.
\begin{table}[t]\scriptsize
\centering
\scriptsize
\caption{Ablation: localization similarity metric and supervision on CelebA-HQ (SD Inpainting).}
\label{tab:loc_similarity_metric}
\renewcommand\arraystretch{1.05}
\resizebox{\columnwidth}{!}{%
\setlength{\tabcolsep}{4mm}{
\begin{tabular}{@{}lccc@{}}
\toprule
Setting (Factor) & F1 & AUC & mIoU \\
\midrule
\multicolumn{4}{l}{\textit{Design}} \\
Seg-only (no similarity branch)               & 0.912 & 0.966 & 0.842 \\
Seg + Similarity (ours)               & \textbf{0.946} & \textbf{0.975} & \textbf{0.941} \\
\midrule
\multicolumn{4}{l}{\textit{Supervision}} \\
Original-only (genuine pairs)                 & 0.881 & 0.939 & 0.808 \\
Orig. + Synthetic (ours)             & \textbf{0.946} & \textbf{0.975} & \textbf{0.941} \\
\bottomrule
\end{tabular}%
}}
\end{table}

\begin{table}[t]
\centering
\caption{Ablation on face-representation guidance. Evaluation on CelebA-HQ under SD Inpainting (1,000 images). \emph{PSNR}$_{\emph{rec}}$ in dB; Bit Acc. denotes watermark extraction accuracy; latency is per-image decode time on one RTX 3090.}
\label{tab:face_repr_ablation}
\setlength{\tabcolsep}{8pt}
\resizebox{\columnwidth}{!}{%
\begin{tabular}{@{}lcccc@{}}
\toprule
Guidance & \emph{PSNR}$_{\emph{rec}}$ & \emph{SSIM}$_{\emph{rec}}$ & Bit Acc. (\%) & Latency (ms) \\
\midrule
256-D embedding               & 23.15 & 0.709 & 97.54    & 49.77 \\
576-D embedding               & 25.83 & 0.843 & 96.06  & 51.15 \\
1024-D embedding              & 31.05 & 0.894 & 97.29  & 53.57 \\
\bottomrule
\end{tabular}}
\end{table}

\noindent\textbf{Ablation on Watermark Guidance for Recovery.}
We evaluate the impact of the dimensionality of the VAE-based facial latent representation on recovery, considering three settings: 256, 576, and 1024 dimensions. 
The latent dimensionality is controlled by resizing the input image to different resolutions before VAE encoding. As shown in Table~\ref{tab:face_repr_ablation}, higher-dimensional latents yield improved PSNR$_{\mathrm{rec}}$ and SSIM$_{\mathrm{rec}}$ with negligible effect on watermark extraction accuracy and inference latency. This indicates that richer semantic priors improve  fidelity without compromising robustness.

\begin{table}[t]
\centering
\caption{Robustness against watermark removal attacks. Bit extraction accuracy (\%) under three representative removal attacks on 1,000 CelebA-HQ images.}
\label{tab:watermark_removal_attacks}
\begin{adjustbox}{max width=\linewidth}
\begin{tabular}{lcccc}
\toprule
Method & \multicolumn{1}{l}{VIC \cite{ball2018variationalimagecompressionscale}} & \multicolumn{1}{l}{DGML \cite{9156817}} & \multicolumn{1}{l}{WMAttacker \cite{zhao2024invisible}} & \multicolumn{1}{l}{Average} \\
\midrule
SepMark \cite{wu2023sepmark} & 91.94 & 89.49 & 68.79 & 83.41 \\
TrustMark \cite{bui2025trustmark} & \textbf{99.98} & \textbf{100.00} & 77.09 & 92.36 \\
EditGuard \cite{zhang2024editguard} & 50.71 & 51.03 & 50.59 & 50.78 \\
OmniGuard \cite{zhang2025omniguard} & 98.81 & 98.73 & 87.56 & \underline{94.97} \\
StableGuard \cite{StableGuard} & 56.63 & 55.12 & 56.67 & 56.14 \\
WAM \cite{sander2025watermark} & 74.81 & 75.31 & \underline{97.84} & 82.66 \\
VeriFi (Ours) & \underline{99.26} & \underline{99.04} & \textbf{99.58} & \textbf{99.29} \\
\bottomrule
\end{tabular}
\end{adjustbox}
\end{table}

\noindent\textbf{Robustness to watermark removal attacks.}
Following~\cite{zhao2024invisible}, we evaluate the robustness of VeriFi against three representative watermark removal attacks: VIC \cite{ball2018variationalimagecompressionscale}, DGML \cite{9156817}, and WMAttacker~\cite{zhao2024invisible}. Table~\ref{tab:watermark_removal_attacks} reports bit extraction accuracy under these attacks on 1,000 CelebA-HQ images. VeriFi achieves the highest average robustness of 99.29\%, outperforming all baselines. This demonstrates that VeriFi's robust embedding and decoding mechanisms effectively resist sophisticated removal attempts, ensuring reliable ownership verification even under adversarial conditions.

\section{Conclusion}
We propose VeriFi, a unified and versatile watermarking framework designed for robust copyright tracing, precise tamper localization, and high-fidelity facial recovery. By jointly embedding ownership signatures and semantic facial representations, VeriFi achieves a synergistic effect where decoded signals serve as powerful priors for both localization and restoration. Extensive experiments on CelebA-HQ and FFHQ benchmarks validate the superiority of our approach in terms of extraction accuracy, localization precision, and recovery quality. Our method offers a comprehensive defense for protecting and verifying digital facial assets against increasingly sophisticated generative models. Future research will explore extending this multi-task framework to video content and other sensitive biometric modalities.

\section*{Acknowledgments}
This work is supported in part by the Guangdong S\&T Program (2026B0101100003), by the National Natural Science Foundation of China (U23B2023 and 62472199), by Guangdong Key Laboratory of Data Security and Privacy Preserving (2023B1212060036), Guangdong Hong Kong Joint Laboratory for Data Security and Privacy Protection (2023B1212120007), by the Basic and Applied Basic Research Foundation of Guangdong Province (2025A1515011097), and by the Outstanding Youth Project of Guangdong Basic and Applied Basic Research Foundation (2023B1515020064). This work is also supported by the Engineering Research Center of Trustworthy AI, Ministry of Education.
\bibliography{main.bib}
\bibliographystyle{ACM-Reference-Format}

\end{document}